\pdfoutput=1

\documentclass[11pt]{article}

\usepackage[]{ACL2023}

\usepackage{times}
\usepackage{latexsym}

\usepackage[T1]{fontenc}

\usepackage[utf8]{inputenc}

\usepackage{microtype}

\usepackage{inconsolata}

\usepackage{graphicx}
\usepackage{multirow}
\usepackage{booktabs}
\usepackage{amssymb}
\usepackage{subfigure}
\usepackage{bbding}
\usepackage{pifont}
\usepackage{amsmath}

\title{KNSE: A Knowledge-aware Natural Language Inference Framework for Dialogue Symptom Status Recognition} 

\author{
Wei Chen$^1$,
Shiqi Wei$^1$,
Zhongyu Wei$^{1,2}$\Thanks{ Corresponding author.}, 
{\bf Xuanjing Huang$^{3}$}\\
\textsuperscript{\rm 1}School of Data Science, Fudan University, China\\
\textsuperscript{\rm 2}Research Institute of Intelligent and Complex Systems, Fudan University, China\\
\textsuperscript{\rm 3}School of Computer Science, Fudan University, China\\
\texttt{\{chenwei18,sqwei19,zywei,xjhuang\}@fudan.edu.cn}
\ \ \ 
}

\begin{document}
\maketitle
\begin{abstract}




Symptom diagnosis in medical conversations aims to correctly extract both symptom entities and their status from the doctor-patient dialogue. In this paper, we propose a novel framework called KNSE for symptom status recognition (SSR), where the SSR is formulated as a natural language inference (NLI) task. For each mentioned symptom in a dialogue window, we first generate knowledge about the symptom and hypothesis about status of the symptom, to form a \emph{(premise, knowledge, hypothesis)} triplet. The BERT model is then used to encode the triplet, which is further processed by modules including utterance aggregation, self-attention, cross-attention, and GRU to predict the symptom status. Benefiting from the NLI formalization, the proposed framework can encode more informative prior knowledge to better localize and track symptom status, which can effectively improve the performance of symptom status recognition. Preliminary experiments on Chinese medical dialogue datasets show that KNSE outperforms previous competitive baselines and has advantages in cross-disease and cross-symptom scenarios.

\end{abstract}




\section{Introduction}

Dialogue symptom diagnosis is an important task in medical dialogue modeling, which is widely used in automatic construction of electronic medical records (EMRs)~\cite{du2019extracting,lin-etal-2019-enhancing,gu2020automatic,zhang-etal-2020-mie} and automatic diagnosis systems~\cite{wei2018task,xu2019end,zhong2022hierarchical,chen2023dxformer}. Dialogue symptom diagnosis can be defined as two subtasks: symptom entity recognition (SER) and symptom status recognition (SSR). The former aims to identify symptom entities from doctor-patient dialogues, while the latter aims to further clarify the relationship between identified symptoms and patients. 

\begin{figure}
\centering
\includegraphics[width=1.0\columnwidth]{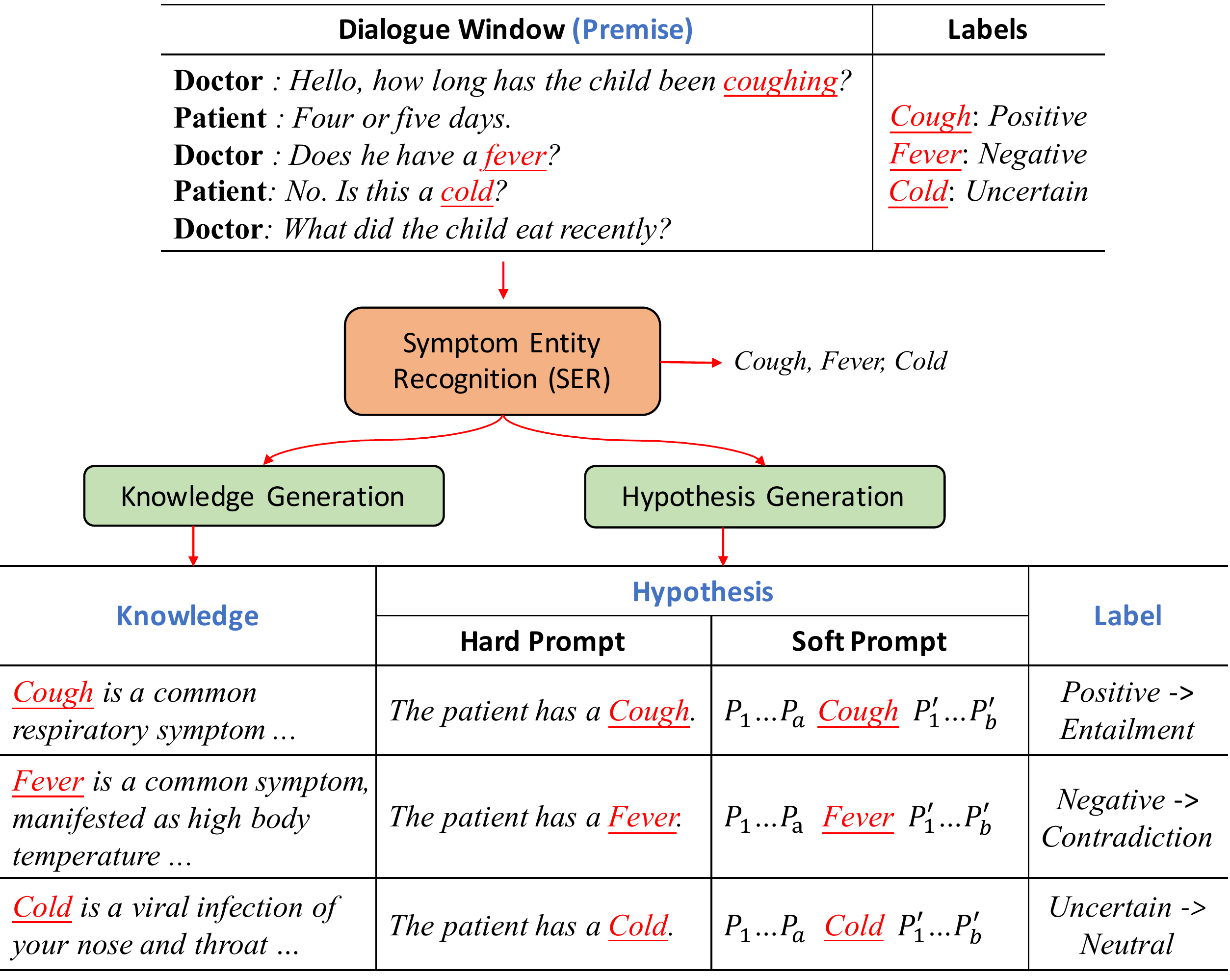}
\caption{Data variations for dialogue symptom diagnosis in our KNSE framework. The Premise, Knowledge, Hypothesis, and Label (in blue font) are the converted data for NLI training.}
\label{fig:arch}
\end{figure}



Diagnosing symptoms from doctor-patient dialogues is challenging due to the common nonstandard oral expressions in dialogues~\cite{lin-etal-2019-enhancing} and the fact that status information of a single symptom entity may be scattered across multiple utterances~\cite{lou2021mlbinet}. Existing studies try to solve these issues by sequence labeling~\cite{finley2018automated,zhao2021knowledge}, generative modeling~\cite{du2019extracting,he2021document}, semantic integration~\cite{lin-etal-2019-enhancing}, context modeling~\cite{zeng2022csdm,hu2022contextual,dai2022chinese}, etc. However, previous studies have the following limitations: 1) in most systems, symptoms and their status are jointly predicted, which makes it difficult to generalize to unseen symptoms; 2) symptom-related knowledge is rarely considered. 





In this paper, we regard symptom status recognition (SSR) as a natural language inference (NLI) task, and propose a novel framework called KNSE for SSR task, as shown in Figure~\ref{fig:arch}. For each symptom mentioned in a given dialogue window, we first generate knowledge about the symptom and hypothesis about the status of symptom, and construct a triplet of the form (premise, knowledge, hypothesis), where the premise is text of the dialog window. After encoding the concatenated text of the triples using BERT, KNSE further utilizes utterance aggregation, self-attention and cross-attention modules to extract knowledge and hypothesis related matching features, and adopts GRU module to track these matching features to generate symptom status features. Preliminary experiments on CMDD and its variant datasets demonstrate that KNSE outperforms previous competing baselines and has advantages in cross-disease and cross-symptom scenarios.

\section{Related Work}


\paragraph{Medical Dialogue Dataset} ~ {Extracting structured information from medical dialogues has received increasing attention, and various human-annotated medical dialogue datasets are constructed to support this research. \citeauthor{du2019extracting} annotated a Chinese dialogue corpus, where 186 symptoms are defined and each symptom is associated with a three-value status (Positive, Negative, Unknown). \citeauthor{lin-etal-2019-enhancing} constructed a chinese medical dialogue datasets called CMDD, where each symptom is annotated in the dialogue with BIO format, with its corresponding status and standardized name. \citeauthor{zhang-etal-2020-mie} created CHYU dataset containing 1,120 dialogues, where more medical entity categories and their status are annotated, including symptoms, tests, operations, etc. \citeauthor{chen2023benchmark} created IMCS-21, a more extensive manually annotated medical conversation dataset, including medical entities and status, dialog intentions, and medical reports.}






\paragraph{Medical Information Extraction} ~ {Several methods have been proposed for extracting structured information from medical dialogues. \citeauthor{finley2018automated} first introduced a linear processing pipeline system to automatically extract EMRs from oral medical dialogues. \citeauthor{du2019extracting} developed a span-attribute tagging model and a Seq2Seq model to infer symptom entities and their status from medical dialogues. \citeauthor{lin-etal-2019-enhancing} utilized attention mechanism and symptom graph to integrate semantic information across sentences. \citeauthor{zeng2022csdm,hu2022contextual,hu2022contextual} proposed context modeling approaches to learn the joint representation of context and symptoms. The closest study to our method for dialogue symptom diagnosis is the machine reading comprehension (MRC) framework  proposed by \citeauthor{zhao2021knowledge}, in which the author adopted a similar sentence pair classification method to identify the status of each symptom. Our method extends this approach by introducing additional symptom knowledge and adopting a more complex network structure compared to a simple BERT~\cite{devlin-etal-2019-bert} encoder.}

\section{Method}

\subsection{Task Formulation}





Given a dialogue window $X = \{U_{1}, U_{2}, ... ,U_{n}\}$, where $U_{i}$ represents a patient (or doctor) utterance, and $n$ is the window size. The set of all symptoms is denote as $T$, and the set of all symptom status is denoted as $S$. 

For each dialogue window $X$, dialogue symptom diagnosis task aims to extracting all mentioned symptoms and corresponding status, i.e., $y = \{(t_{i}, s_{i})\}_{i=1}^{k}$, where $t_{i} \in T$, $s_{i} \in S$.

\subsection{Symptom Entity Recognition}



Since symptom entity recognition (SER) is not the focus of this study, we simply utilize the BERT-CRF~\cite{devlin-etal-2019-bert} model to extract the text span of symptom entities, and adopt a SVM~\cite{hearst1998support} classifier to standardize the identified symptom entities. The BERT-CRF and SVM models achieved 95\% F1 scores and 98\% accuracy on the test set respectively, indicating the SER task is relatively simple and will not bring too much noise to the next step, which is consistent with the conclusion in \cite{zhao2021knowledge}. Therefore, improving the performance of symptom status recognition (SSR) is currently the most pressing obstacle. We will focus on the major contributions of the proposed KNSE framework in subsequent chapters.

\subsection{KNSE Framework}



\subsubsection{Symptom Hypothesis Generation}






We regard SSR task as a natural language inference (NLI)~\cite{chen-etal-2017-enhanced} problem, where the concatenated text of dialogue window is regarded as the \emph{premise}, and the statement of symptom status is regarded as the \emph{hypothesis}. We set the hypothesis that \textbf{the patient has the given symptom} (Figure \ref{fig:arch}), and consider two ways to generate the hypothesis, i.e., hard prompt and soft prompt. 



\paragraph{Hard Prompt} ~ {The hard prompt based template is set to "\emph{The patient has a \{\}.}", where the content in curly brackets is filled with the given symptom.}




\paragraph{Soft Prompt} ~ {The soft prompt template is set to "$P_{1}...P_{a}\;\{\}\;P_{1}^{'}...P_{b}^{'}$", where $a$ and $b$ prompt tokens are added before and after the given symptom respectively, and the embeddings of these prompt tokens are trainable. Note that $a$ and $b$ are hyperparameters.}




\subsubsection{Symptom Knowledge Generation}

Knowledge about symptoms may help better localize positive symptoms, we utilize large language models (LLMs) to obtain symptom knowledge for convenience. We first construct the following question template, "\emph{Please briefly describe the \{\} symptom}", where the content in curly brackets will be filled with a specific symptom. Then we feed the question to ChatGPT~\cite{chatgpt}, and cache the answer as the symptom knowledge. It is worth noting that the acquisition of symptom knowledge does not rely on LLMs, as it can be obtained through other sources, such as relevant entries on Wikipedia, professional medical websites, etc.

\subsubsection{Natural Language Inference}


For a given triplet $(P, H, K)$, i.e., the premise $P$, the generated hypothesis $H$ and knowledge $K$, the natural language inference (NLI) module aims to predict whether the hypothesis is true (entailment), false (contradiction), or undetermined (neutral), given the premise and the knowledge. Inspired by recent studies in multi-turn dialogue modeling~\cite{zhang2018modeling,chen2022contextual,chen2022dialogved}, we propose a similar matching model, which consists of modules including encoder, utterance aggregation, self attention, cross attention and GRU~\cite{dey2017gate}. 



\paragraph{Encoder} ~ {We first adopt BERT~\cite{devlin-etal-2019-bert} to encode the triplet. We concatenate the triplet with special token [SEP] and feed them into BERT to obtain their respective hidden vectors $H_{P}$, $H_{H}$ and $H_{K}$, whose dimension is the corresponding length multiplied by the hidden vector dimension $d$: 
$$H_{P}, H_{H}, H_{K} = {\rm Encoder}(P, H, K)$$
}
 






\paragraph{Utterance Aggregation} ~ {The hidden vector of hypothesis $H_{H}$ is then treated as a query to apply an attention mechanism to the hidden vector of premise $H_{P}$, which can aggregate the hypothesis related information from each utterance into a single vector. The process of aggregation works as: 
\begin{equation}
\begin{aligned}
    a[i][j] &= \max_{k}(H_{P}[i][j] W H_{H}[k]^{T}) \\ 
    p[i] &= {\rm softmax}(a[i]) \\
    C_{hyp}[i] &= \sum_{j} p[i][j]H_{P}[i][j],  
\end{aligned}    
\end{equation}
where $W \in \mathcal{R}^{d\times d}$ is trainable weights, $[i][j]$ represents the $j$-th word in the $i$-th utterance of premise, $[k]$ represents the $k$-th word in the hypothesis, and $C_{hyp}[i]$ is the aggregated hypothesis-related information of the $i$-th utterance of the premise. The process will assign high values to the words related to the hypothesis, and thus extract the most relevant information within an utterance. }

\begin{figure}
\centering
\includegraphics[width=1.0\columnwidth]{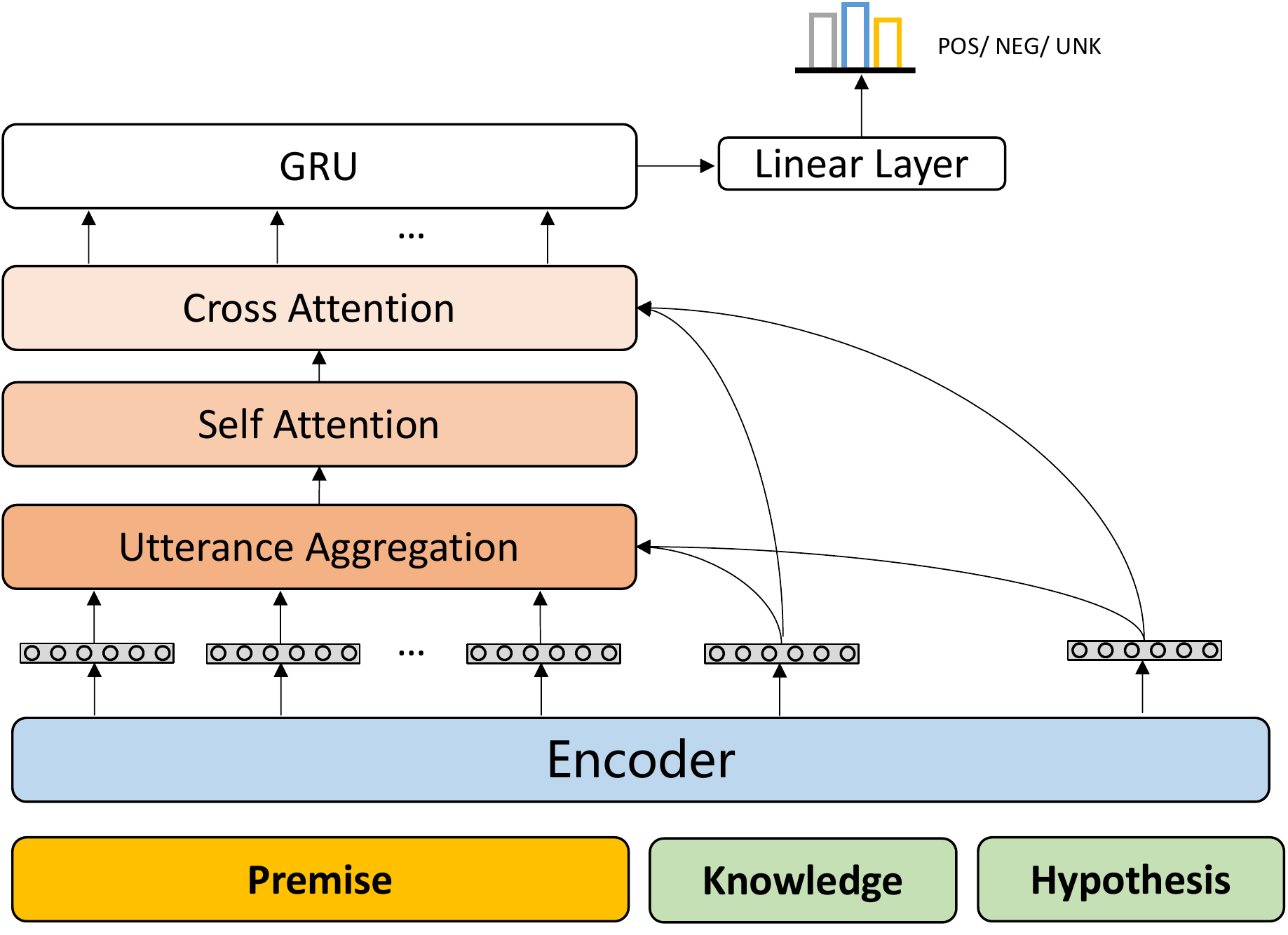}
\caption{The structure of natural language inference module.}
\label{fig:model}
\end{figure}


\paragraph{Self Attention} ~ {We use self-attention to enhance contextual utterance representation as follows:  
\begin{equation}
\begin{aligned}
C_{hyp}^{sa} &= {\rm SA}(C_{hyp}, C_{hyp}, C_{hyp}), 
\end{aligned}   
\end{equation}
where $\rm SA(\cdot)$ represents a combination of self-attention, residual connection, and layer normalization modules.}


\paragraph{Cross Attention} ~ {We adopt cross-attention to enhance hypothesis information fusion as follows: 
\begin{equation}
\begin{aligned}
C_{hyp}^{ca} &= {\rm CA}(C_{hyp}^{sa}, H_{H}, H_{H}),  
\end{aligned}   
\end{equation}
where $\rm CA(\cdot)$ represents a combination of cross-attention, residual connection, and layer normalization modules.}






\paragraph{GRU} ~ {We have presented $C_{hyp}^{ca}$, which is regarded as the matching features between hypothesis and premise. In the same way, we can obtain $C_{knw}^{ca}$, i.e., the matching feature between knowledge and premise, by using knowledge as the key in utterance aggregation module. Afterwards, we concatenate $C_{hyp}^{ca}$ and $C_{knw}^{ca}$ and update the matching features using a bidirectional GRU~\cite{dey2017gate} as follows:
\begin{equation}
\begin{aligned}
\hat{h} &= {\rm GRU}([C_{P}^{ca}; C_{K}^{ca}]),
\end{aligned}  
\end{equation}
where $\hat{h}$ is our final representation of symptom status, which is recursively updated from sentence-level matching features. We employ a linear layer to map $\hat{h}$ to the probability distribution of symptom status and train KNSE with the cross entropy objective.}

\begin{table*}
\small
\centering
\begin{tabular}{lcccccc} \toprule
\multirow{2}{*}{\textbf{Model}} & \multicolumn{3}{c}{\textbf{Window-Level}} & \multicolumn{3}{c} {\textbf{Dialogue-Level}} \\
                       & \textbf{Precision}         & \textbf{Recall}         & \textbf{F1 score}       & \textbf{Precision}          & \textbf{Recall}         & \textbf{F1 score}        \\ \midrule
Plain-Classifier~\cite{zhou2016attention}        & 79.80      & 75.90      & 76.84    & 67.81      & 67.81     & 65.58     \\
BERT-MTL~\cite{devlin-etal-2019-bert}        & 80.20      & 77.18      & 77.25    & 70.28      & 70.12     & 68.21     \\
MIE-multi~\cite{zhang-etal-2020-mie}              & 81.63     & 80.45     & 80.23    & 74.12      & 72.38     & 72.51     \\
MRC~\cite{zhao2021knowledge}           & 80.05    & 78.45      & 79.24    & 73.56      & 74.92     & 74.24     \\
CANE~\cite{hu2022contextual}                   & 82.54     & 81.36     & 81.33    & 75.78      & 75.79     & 75.20      \\  \midrule
\textbf{KNSE (Ours)}                           & \textbf{84.17}     & \textbf{82.86}    & \textbf{83.57}         & \textbf{77.32}           & \textbf{76.59}          & \textbf{76.83}           \\  \midrule
\multicolumn{7}{l}{\textbf{Ablation Study}}  \\  \midrule
\qquad encoder only    &  82.12         & 80.74          & 80.84          & 74.81           & 74.59          & 74.52          \\ 
\qquad hard prompt    &  83.78         & 82.52          & 83.14          & 76.92           & 76.21          & 76.39          \\
\qquad w/o. knowledge          &  83.05         & 82.40          & 82.93          & 75.89           & 76.03          & 75.89   \\ \bottomrule        
\end{tabular}
\caption{Experimental results on CMDD dataset.}
\label{tab:results1}
\end{table*} 

\section{Experiments}

\subsection{Dataset} 

We conduct extensive experiments on Chinese Medical Diagnosis Dataset (CMDD)~\cite{lin-etal-2019-enhancing} to demonstrate the effectiveness of our framework on dialogue symptom diagnosis task. We convert CMDD to a sliding window format on all dialogues, with a windows size of 5, following the settings of previous studies~\cite{hu2022contextual}.


The CMDD dataset contains 2,067 dialogues and 87,005 windows, including 52,952/16,935/17,118 dialogue windows in train/develop/test sets, respectively, covering 160 symptoms, and 3 possible status (Positive, Negative, and Unknown) for each symptom.

\subsection{Baselines}

Five baseline models are used for comparison, including Plain-Classifier~\cite{zhou2016attention}, BERT-MTL~\cite{devlin-etal-2019-bert}, MIE-multi~\cite{zhang-etal-2020-mie}, MRC~\cite{zhao2021knowledge} and CANE~\cite{hu2022contextual}. The Plain-Classifier, MIE-multi and CANE models regard the task as multi-label classification problem and jointly predict the symptoms and their status; while BERT-MTL and MRC models adopt a pipeline approach, i.e., first predict the mentioned symptoms, and then predict the status of each symptom.

We also compare several variants of KNSE. Encoder only represents after encoding the triplet, the hidden vector of [CLS] is directly used to predict the symptom status. Hard prompt indicates using fixed, non-trainable prompt to generate the hypothesis. KNSE w/o. knowledge means not using knowledge. 



\subsection{Experimental Settings}
\label{sec:settings}

We use BERT~\cite{devlin-etal-2019-bert} as our encoder. We set the maximum length of each utterance to 50 to ensure that the length of the dialogue window does not exceed 256, and we set the maximum length of symptom knowledge to 64. We use the AdamW~\cite{loshchilov2017decoupled} as the optimizer, and its betas, weight decay and other parameters follow the settings in RoBERTa~\cite{liu2019roberta}. We set the batch size as 64, the learning rate as 1e-5. We adopt soft prompt and the hyper-parameter $a$ and $b$ are set to 10 and 5 respectively. We train a total of 20 epochs and choose the model that performs best on the develop set. 

 
\subsection{Evaluation Metrics}

We report the micro-averaged Precision, Recall and F1 score in the multi-label classification~\cite{zhang2013review} scenario to measure the overall performance of the system, where the label space is $|T|*|S|$, and only if both the symptom and its status are correct can they be considered as real positive cases. Both the window-level and the dialogue-level results are reported in the paper, see details in \cite{hu2022contextual}.

\subsection{Main Results}

Table \ref{tab:results1} shows the experimental results on the CMDD dataset. It can be seen that KNSE outperforms all baselines in both window-level and dialogue-level evaluation metrics. This illustrates the effectiveness of the KNSE framework. It is worth noting that since each symptom is identified independently, KNSE does not take advantage of the co-occurrence of some symptoms and their status like MIE-multi and CANE. The results in window-level are relatively higher than the results in dialogue-level. This is because the latter is stricter than the former, which has been verified in previous studies~\cite{hu2022contextual}. 

It is more interesting to analyze the effectiveness of KNSE components. From the results of the ablation experiments: The variant KNSE encoder only underperforms, suggesting that the inductive bias introduced by these additional modules in addition to the encoder are effective for symptom status representation learning; Using hard prompt tokens instead of soft prompt tokens will slightly reduce the model performance, we guess that it may be because tunable soft prompts can help the model learn to pay attention to important words in the dialogue window; Introducing symptom knowledge is effective, since intuitively, knowledge can help us better identify positive symptoms.

\begin{table}
\centering
\small
\begin{tabular}{llccc} \toprule
\textbf{Model}               & \textbf{Dataset}       & \textbf{POS} & \textbf{NEG} & \textbf{UNK} \\ \midrule
\multirow{3}{*}{MRC(~\citeauthor{zhao2021knowledge})} & CMDD    &  87.1       &  73.8        & 72.4                 \\
                      & CD-CMDD                 &  81.8       &  64.3     & 63.2                 \\ 
                      & CS-CMDD                 & 74.2        &  58.7    & 56.6                 \\ \midrule
\multirow{3}{*}{\textbf{KNSE (Ours)}} & CMDD    &  89.6       &  76.5    & 74.2             \\
                      & CD-CMDD                 &  86.5       &  72.7     & 71.4                 \\ 
                      & CS-CMDD                 &  83.4        & 69.8     & 66.5            \\ \bottomrule
\end{tabular}
\caption{Experimental results on cross-domain variants of CMDD dataset, where POS, NEG and UNK are the abbreviations of Positive, Negative and Unknown respectively.} 
\label{tab:results2}
\end{table}



\subsection{Cross-domain Scenarios}

We further explore the model performance in cross-disease and cross-symptom scenarios. Specifically, we redivide the CMDD dataset, where CMDD-CD and CMDD-CS represent the division of training sets, develop sets and test sets according to disease and symptom, respectively. In CMDD-CD dataset, the diseases in the test set are not seen in the training set, but there may be some overlapping symptoms. In CMDD-CS dataset, all symptoms in the test set in do not appear in the training set. 


We assume that the symptoms are already known, and adopt the MRC and KNSE models to predict the status of each symptom. We report F1 score for each category of symptom status in Table~\ref{tab:results2}. The experimental results show that the dataset divided by symptoms is more difficult on the SSR task than the dataset divided by diseases, which is intuitive. Besides, it can be seen that the performance degradation of KNSE on CD-CMDD and CS-CMDD datasets (about 3\textasciitilde8\%) is much lower than that of MRC (about 9\textasciitilde16\%), suggesting that compared with MRC, KNSE has a stronger ability to recognize the status of symptoms that have not been seen in the training set. 





\section{Conclusion}

In this paper, we investigate the problem of symptom diagnosis in doctor-patient dialogues. We proposed a knowledge-aware framework by formalizing the symptom status recognition problem as a natural language inference task. Our framework is able to encode more informative prior knowledge to better localize and track symptom status, which can effectively improve the performance of symptom status recognition. We develop several competitive baselines for comparison and conduct extensive experiments on the CMDD dataset. The experimental results demonstrate the effectiveness of our framework, especially in cross-disease and cross-symptom scenarios. 








\section*{Acknowledgments}

This work is supported by National Natural Science Foundation of China (No. 6217020551) and Science and Technology Commission of Shanghai Municipality Grant (No.21QA1400600).

\bibliography{custom}
\bibliographystyle{acl_natbib}

\appendix

\end{document}